\documentclass[11pt]{article}

\usepackage[preprint]{acl} 

\usepackage[T1]{fontenc}
\usepackage[utf8]{inputenc}
\usepackage{times}
\usepackage{latexsym}
\usepackage{microtype}
\usepackage{inconsolata}

\usepackage{amsmath}
\usepackage{amssymb}
\usepackage{bm}          

\usepackage{graphicx}

\usepackage{booktabs}
\usepackage{multirow}
\usepackage{array}

\usepackage{algorithm}
\usepackage{algorithmic}

\usepackage{hyperref}
\hypersetup{
  colorlinks=true,
  linkcolor=blue,
  citecolor=blue,
  urlcolor=blue
}

\usepackage{cleveref}

\usepackage{enumitem}

\title{Knowledge Reasoning of Large Language Models Integrating Graph-Structured Information for Pest and Disease Control in Tobacco}

\author{Siyu Li,  Chenwei Song, Wan Zhou, Xinyi Liu \\
School of Information Science and Engineering\\ Chongqing Jiaotong University \\
\{lsy20, xyliu\}@mails.cqjtu.edu.cn 
}

\begin{document}
\maketitle
\begin{abstract}
This paper proposes a large language model (LLM) approach that integrates graph-structured information for knowledge reasoning in tobacco pest and disease control. Built upon the GraphRAG framework, the proposed method enhances knowledge retrieval and reasoning by explicitly incorporating structured information from a domain-specific knowledge graph. Specifically, LLMs are first leveraged to assist in the construction of a tobacco pest and disease knowledge graph, which organizes key entities such as diseases, symptoms, control methods, and their relationships. Based on this graph, relevant knowledge is retrieved and integrated into the reasoning process to support accurate answer generation.
The Transformer architecture is adopted as the core inference model, while a graph neural network (GNN) is employed to learn expressive node representations that capture both local and global relational information within the knowledge graph. A ChatGLM-based model serves as the backbone LLM and is fine-tuned using LoRA to achieve parameter-efficient adaptation. Extensive experimental results demonstrate that the proposed approach consistently outperforms baseline methods across multiple evaluation metrics, significantly improving both the accuracy and depth of reasoning, particularly in complex multi-hop and comparative reasoning scenarios.
\end{abstract}

\section{Introduction}

Tobacco is an important economic crop in southern China and is highly susceptible to pest and disease infestations during its growth process. Tobacco pests and diseases not only severely affect yield and quality, but can also cause significant economic losses. Therefore, pest and disease control in tobacco agriculture has long relied on expert experience and observational judgment, which often suffer from low efficiency and high error rates. With the rapid development of information technology, artificial intelligence has shown broad application prospects in agriculture \citep{kamilaris2018deep,li2021survey}. Knowledge reasoning based on large language models (LLMs) has gradually emerged as a new paradigm in the agricultural domain \citep{yang2023llm}.

Compared with traditional agricultural knowledge organization methods, knowledge graphs can more systematically store and manage dispersed agricultural knowledge by constructing explicit networks of entities and relations, thereby providing a solid foundation for intelligent agricultural applications \citep{paulheim2017knowledge}. Meanwhile, recent breakthroughs in natural language understanding by LLMs have enabled more effective language-based reasoning and interaction \citep{brown2020gpt3,wei2022chain}. As a result, combining the structured information of knowledge graphs with the semantic reasoning capabilities of LLMs has become an active research direction \citep{lewis2020rag,yasunaga2021qagnn}. In the context of tobacco pest and disease control, integrating agricultural knowledge graphs with LLMs can support tasks such as disease diagnosis and control strategy recommendation by reasoning over complex information, including pests, diseases, control methods, and crop growth conditions.

However, improving the reasoning performance of LLMs in this domain remains a challenge. Existing approaches mainly rely on prompt engineering or simple knowledge injection, often neglecting the rich relational information between entities, which limits the reasoning capability of LLMs to some extent \citep{liu2023survey}. To address this issue, this paper proposes a large language model approach that integrates graph-structured information for domain-specific knowledge reasoning. Built upon the GraphRAG framework \citep{edge2024graphrag}, graph embeddings (TransE) \citep{bordes2013transe} and graph neural networks (GCN) \citep{kipf2017semi} are introduced to enhance node representation learning, thereby improving the reasoning process of LLMs and the accuracy of the question-answering system.

Compared with existing studies, the innovations and advantages of this work lie in a more comprehensive utilization of knowledge graph information. The proposed method not only leverages entity-level information, but more importantly incorporates relational structure through graph embedding and graph neural network modeling, providing LLMs with richer contextual knowledge and stronger reasoning capability. In addition, this paper presents a novel graph--LLM integration approach tailored for tobacco pest and disease control. By integrating GCN-learned graph embeddings into the input or intermediate representations of LLMs, the proposed framework enables deeper fusion between structured knowledge and language models, allowing more effective utilization of the relational information encoded in knowledge graphs.

\section{Related Work}

\paragraph{Knowledge Graphs for Agricultural Pest and Disease Control.}
As an effective knowledge representation paradigm, knowledge graphs have in recent years been widely applied in agricultural pest and disease control \citep{paulheim2017knowledge}. By integrating heterogeneous agricultural knowledge and constructing explicit networks of entities and relations, knowledge graphs enable more systematic organization and management of domain knowledge. In the field of crop pest and disease knowledge graphs, typical entities include disease names, symptoms, control methods, and crop varieties \citep{li2021survey}. In addition, domain-specific pest and disease knowledge graphs have been constructed for particular crops such as tobacco and wheat, providing valuable references for agricultural decision support and disease management \citep{kamilaris2018deep}. 
Recent studies have further extended knowledge graph reasoning to temporal settings, enabling models to capture the dynamic evolution of entities and relations over time, which is particularly relevant for modeling disease progression and seasonal pest outbreaks in agriculture \citep{xiongtilp, xiong2024teilp}.

\paragraph{Large Language Models and Graph-Enhanced Reasoning.}
Large language models (LLMs) have demonstrated strong performance in a wide range of natural language processing tasks. Representative models such as GPT-4 and LLaMA exhibit clear advantages in text generation, dialogue, and reasoning, particularly in knowledge understanding and semantic representation \citep{brown2020gpt3,wei2022chain}. However, how to effectively utilize structured knowledge, especially the relational structure encoded in knowledge graphs, remains an important research challenge \citep{yasunaga2021qagnn}.

Graph-based Retrieval-Augmented Generation (GraphRAG) has emerged as an effective paradigm for integrating graph-structured information with language models \citep{edge2024graphrag}. This framework enhances reasoning by learning representations of entities and relations through graph embeddings and graph neural networks, and by integrating these representations into the input or intermediate layers of language models. 
Prior studies have explored the use of graph convolutional networks (GCNs) \citep{kipf2017semi} and graph attention networks (GATs) to learn node representations and incorporate them into the contextual representations of language models, thereby guiding LLMs to reason over relational information in knowledge graphs \citep{yasunaga2021qagnn}. 
These approaches have been shown to significantly improve the performance of language models on knowledge-intensive tasks such as question answering \citep{lewis2020rag}.
Beyond GraphRAG, a growing body of work has explored tighter integration of large language models with graph-structured knowledge, including jointly reasoning over textual representations and knowledge graphs for question answering and commonsense reasoning \citep{yang2024harnessing, xiong2024large, xiong2025deliberate}.

\paragraph{Knowledge Graphs and LLMs in Agricultural Domains.}
At present, research that combines knowledge graphs and LLMs for tobacco pest and disease control remains limited. Existing studies mainly focus on knowledge representation for tobacco pest and disease management, or employ traditional machine learning and deep learning methods for disease recognition and diagnosis \citep{li2021survey}. Some recent work has begun to explore the use of LLMs to enhance agricultural knowledge understanding and management \citep{yang2023llm}, but research that systematically integrates GraphRAG with LLMs for structured reasoning in tobacco pest and disease control is still scarce. Therefore, this paper investigates how to combine the GraphRAG framework with large language models to effectively leverage structured knowledge graph information, with the goal of enhancing the reasoning capability of LLMs in tobacco pest and disease control.

\section{Methodology}

This paper proposes a large language model (LLM) approach that integrates graph-structured information for knowledge reasoning in tobacco pest and disease control. The proposed method is built upon the GraphRAG framework, which incorporates structured knowledge graph information into the reasoning process of LLMs to improve the performance of question-answering systems. The core components of the method include knowledge graph construction, graph embedding learning, graph neural network–based node representation learning, and the fusion of graph representations with LLMs.

\subsection{Knowledge Graph Construction}

First, a tobacco pest and disease knowledge graph is constructed to organize domain knowledge. This graph consists of entities and relations extracted from agricultural literature and expert knowledge. Typical entities include diseases (e.g., \emph{tobacco mosaic disease}), control methods (e.g., \emph{spraying antiviral agents}), and symptoms. Relations describe interactions between entities, such as \emph{treatment-of} relationships between control methods and diseases.

Formally, the knowledge graph is represented as a directed graph
\[
G = (V, E),
\]
where $V$ denotes the set of entities and $E$ denotes the set of relations. Each fact in the graph is represented as an RDF-style triple $(h, r, t)$, where $h$ is the head entity, $r$ is the relation, and $t$ is the tail entity. For example, the triple
\texttt{(tobacco mosaic disease, treated by, spraying antiviral agents)}
indicates that spraying antiviral agents is a treatment method for tobacco mosaic disease.

\subsection{Graph Embedding with TransE}

To encode entities and relations into continuous vector spaces, the TransE model is employed. TransE assumes that relations can be interpreted as translations operating on entity embeddings. Specifically, for a valid triple $(h, r, t)$, the embeddings should satisfy
\[
\mathbf{h} + \mathbf{r} \approx \mathbf{t}.
\]

The training objective of TransE is defined as a margin-based ranking loss:
\begin{equation}\notag
\mathcal{L} =
\sum_{(h,r,t)\in S}
\sum_{(h',r,t')\in S'}
\ell(h,r,t,h',t'),
\end{equation}
\begin{equation}\notag
\ell = 
\Big[
\gamma + d(\mathbf{h}+\mathbf{r}, \mathbf{t})
- d(\mathbf{h'}+\mathbf{r}, \mathbf{t'})
\Big]_+ .
\end{equation}
where $S$ is the set of positive triples, $S'$ is the set of negative triples generated by randomly replacing the head or tail entity, $\gamma$ is a margin hyperparameter, $d(\cdot)$ denotes a distance function (e.g., L1 or L2 norm), and $[\cdot]_+$ denotes the hinge loss.

Through TransE training, each entity and relation in the knowledge graph is mapped to a low-dimensional vector representation that captures relational semantics.

\subsection{Node Representation Learning with GCN}

While TransE captures pairwise relational information, it does not fully exploit the global graph structure. Therefore, a Graph Convolutional Network (GCN) is further applied to refine node representations by aggregating neighborhood information.

Let $\mathbf{h}_i^{(l)}$ denote the representation of node $i$ at the $l$-th GCN layer. The update rule of GCN is given by:
\[
\mathbf{h}_i^{(l+1)} = \sigma \left( \sum_{j \in \mathcal{N}(i)} 
\frac{1}{\sqrt{\deg(i)\deg(j)}} \mathbf{W}^{(l)} \mathbf{h}_j^{(l)} \right),
\]
where $\mathcal{N}(i)$ denotes the set of neighboring nodes of $i$, $\deg(i)$ is the degree of node $i$, $\mathbf{W}^{(l)}$ is a trainable weight matrix, and $\sigma(\cdot)$ is an activation function such as ReLU. The normalization term $\frac{1}{\sqrt{\deg(i)\deg(j)}}$ is used to stabilize training and prevent numerical explosion.

By stacking multiple GCN layers, each node can aggregate information from higher-order neighborhoods. For example, the representation of the disease node \emph{tobacco mosaic disease} can incorporate features from related control methods and symptoms, resulting in a more informative embedding.

\subsection{Integration with Large Language Models}

On the LLM side, modern Transformer-based models such as LLaMA and ChatGLM are adopted as the backbone. Given an input token sequence
\[
X = (x_1, x_2, \ldots, x_n),
\]
the LLM models the conditional probability of the output sequence as:
\[
P(X) = \prod_{i=1}^{n} p(x_i \mid x_{<i}).
\]

Transformer-based LLMs rely on the self-attention mechanism, which is computed as:
\[
\text{Attention}(Q, K, V) = \text{softmax}\left(\frac{QK^\top}{\sqrt{d_k}}\right)V,
\]
where $Q$, $K$, and $V$ are the query, key, and value matrices, respectively, and $d_k$ is the dimensionality of the key vectors. Positional encoding is applied to preserve the order information of tokens in the sequence.

\subsection{GraphRAG-Based Fusion}

The GraphRAG framework retrieves graph information relevant to a given query and integrates it into the LLM input. For a query such as \emph{``How to prevent tobacco mosaic disease?''}, the system first identifies the corresponding disease entity in the knowledge graph. The TransE embedding $\mathbf{e}_d \in \mathbb{R}^k$ of the disease entity is then refined through GCN to obtain $\mathbf{g}_d \in \mathbb{R}^k$.

The query embedding and graph embedding are concatenated to form an augmented input:
\[
\mathbf{g}_{\text{input}} = [\mathbf{e}_d; \mathbf{g}_d],
\]
which is fed into the LLM as additional contextual information. Through this mechanism, the LLM is guided to attend to graph-relevant entities such as \emph{spraying antiviral agents} or \emph{field sanitation}, enabling more accurate and comprehensive answers.

\subsection{Summary}

By tightly integrating graph-structured knowledge with large language models, the proposed method effectively combines symbolic relational information and neural language understanding. This fusion significantly enhances the reasoning capability of LLMs in tobacco pest and disease control, leading to more reliable and knowledge-grounded decision support.

\section{Experiments}

To verify the effectiveness of the proposed large language model method integrating graph-structured information for tobacco pest and disease control, extensive experiments were conducted. The experiments focus on evaluating the impact of entity--relation extraction, graph embedding, and graph--LLM fusion on downstream question-answering and reasoning tasks, following common evaluation protocols in knowledge graph--enhanced language modeling \citep{lewis2020rag,yasunaga2021qagnn}.

\begin{table*}[t]
\centering
\small
\caption{Comparison of Different Methods on Knowledge Reasoning Tasks for Tobacco Pest and Disease Control}
\label{tab:comparison}
\begin{tabular}{lcccc}
\toprule
\textbf{Model} &
\textbf{Accuracy (\%)} &
\textbf{Precision (\%)} &
\textbf{Recall (\%)} &
\textbf{F1-score (\%)} \\
\midrule
ChatGLM                & 75.2 & 78.5 & 72.1 & 75.2 \\
KGE + ChatGLM          & 82.5 & 85.3 & 79.8 & 82.4 \\
RAG + ChatGLM          & 85.7 & 87.9 & 83.2 & 85.5 \\
GraphRAG + ChatGLM     & \textbf{90.1} & \textbf{92.3} & \textbf{88.2} & \textbf{90.2} \\
\bottomrule
\end{tabular}
\end{table*}

\subsection{Experimental Setup}

We first constructed a tobacco pest and disease knowledge graph by extracting entities and relations from multiple data sources, including agricultural expert knowledge bases, disease management manuals, and academic literature. The extracted entities cover common tobacco diseases, control methods, pesticide components, symptoms, and causal factors, while relations describe interactions such as \emph{treatment-of} and \emph{prevention-of}. This construction process follows established practices in agricultural knowledge graph development \citep{kamilaris2018deep,li2021survey}. In total, the constructed knowledge graph contains over 1{,}000 entities and relations.

Based on this knowledge graph, we built a tobacco pest and disease question-answering dataset consisting of two parts: (1) a training set used for model learning and (2) a test set used for evaluation. The dataset includes three types of tasks: direct question answering, multi-hop reasoning, and comparative reasoning, which are commonly used to evaluate reasoning capability in knowledge-intensive QA settings \citep{yasunaga2021qagnn,wei2022chain}.

For graph representation learning, the TransE model was adopted to obtain initial embeddings of entities and relations \citep{bordes2013transe}, with an embedding dimension of 100 and a learning rate of 0.01. A Graph Convolutional Network (GCN) was then applied to further refine node representations \citep{kipf2017semi}, using two convolutional layers and ReLU as the activation function.

For the language model component, ChatGLM was selected as the base LLM. To reduce training cost, LoRA was employed for parameter-efficient fine-tuning \citep{hu2022lora}, with the LoRA rank set to 16. Within the GraphRAG framework \citep{edge2024graphrag}, the graph embeddings produced by the GCN were concatenated with text embeddings generated by a Sentence-BERT encoder \citep{reimers2019sentencebert} and then fed into ChatGLM as augmented contextual input.

\subsection{Baselines and Evaluation Metrics}

To comprehensively evaluate the performance of the proposed method, we compared it with the following baselines, which are commonly adopted in knowledge-enhanced language modeling and retrieval-augmented generation research \citep{lewis2020rag,liu2023survey}:

\begin{itemize}
  \item \textbf{ChatGLM}: Direct question answering using ChatGLM without any external knowledge.
  \item \textbf{TransE + LLM}: Using TransE embeddings to retrieve relevant entities and inject them into ChatGLM.
  \item \textbf{KGE + ChatGLM}: A knowledge graph embedding--based method that incorporates entity embeddings into the LLM input.
  \item \textbf{RAG}: A standard retrieval-augmented generation method based on textual retrieval.
\end{itemize}

We adopted Accuracy, Precision, Recall, and F1-score as evaluation metrics to assess performance across different tasks, following standard evaluation practices for classification and question-answering systems \citep{li2021survey}.

\subsection{Results and Analysis}

Experimental results show that the proposed GraphRAG+ChatGLM method consistently outperforms all baseline methods across all evaluation metrics. Compared with using ChatGLM alone, the integration of graph-structured information leads to significant performance improvements, demonstrating the effectiveness of combining knowledge graphs with large language models for domain-specific reasoning tasks.

In particular, the introduction of GCN enables effective aggregation of neighborhood information for each entity, allowing the model to capture complex relational patterns among diseases, symptoms, and control methods. Compared with KGE+ChatGLM, the proposed method achieves notably better performance, highlighting the advantage of explicitly modeling graph structure rather than relying solely on flat entity embeddings.

Compared with the RAG baseline, GraphRAG+ChatGLM shows superior performance in understanding query context and retrieving relevant knowledge. While RAG mainly relies on textual similarity and contextual relevance, it may fail to capture structured relational information. In contrast, GraphRAG leverages the explicit structure of the knowledge graph through GCN-based aggregation, enabling more accurate reasoning and reducing errors caused by irrelevant or redundant retrieved information.

The advantages of GraphRAG+ChatGLM are particularly evident in multi-hop and comparative reasoning tasks. For example, in questions such as \emph{``Which pesticides are suitable for both tobacco mosaic disease and aphid control?''}, the proposed method can effectively integrate knowledge from different disease entities and identify shared control methods. Similarly, for comparative questions like \emph{``Which control strategy is more effective for biological versus chemical disease prevention?''}, GraphRAG+ChatGLM can reason over relational paths in the knowledge graph to produce more comprehensive and accurate answers.

\subsection{Discussion}

These results demonstrate that GraphRAG+ChatGLM is effective for knowledge reasoning in tobacco pest and disease control, particularly in complex reasoning scenarios. By explicitly incorporating graph structure, the proposed method improves both the depth and accuracy of reasoning. The GCN component plays a crucial role in enhancing entity representations, while the LLM benefits from structured contextual guidance.

However, the current study has several limitations. The scale of the constructed dataset is relatively small, and future work will focus on building larger and more comprehensive knowledge graphs to further validate the effectiveness and generalizability of the proposed approach. In addition, future research may explore more advanced graph neural network architectures and more efficient fusion strategies to further enhance the reasoning capability of large language models in agricultural domains.

\section{Conclusion}

This paper proposes a large language model approach that integrates graph-structured information for knowledge reasoning in tobacco pest and disease control. Built upon the GraphRAG framework, the proposed method incorporates graph neural networks (GCNs) to inject structured knowledge graph information into ChatGLM. Experimental results demonstrate that the proposed approach consistently outperforms baseline methods across multiple evaluation metrics, with particularly notable improvements in multi-hop and comparative reasoning tasks. These results validate the effectiveness of integrating graph-structured knowledge to enhance the reasoning capability of large language models in agricultural decision-making scenarios.

\section*{Limitations}

Despite the promising results, this study has several limitations. The scale of the constructed tobacco pest and disease knowledge graph is relatively limited, which may restrict the generalization ability of the proposed method. In addition, the current framework relies on fixed graph embeddings and a specific graph neural network architecture, leaving room for further optimization. In future work, we plan to construct larger and more comprehensive knowledge graphs, and to explore more advanced graph representation learning and fusion strategies. We also aim to extend the proposed framework to other knowledge-intensive domains, such as healthcare and intelligent decision-making systems, to further evaluate its general applicability.

\bibliography{custom}

\end{document}